\pdfoutput=1

\documentclass[11pt]{article}

\usepackage[final]{acl}
\usepackage{adjustbox}
\usepackage{pgf}
\usepackage[utf8]{inputenc}
\usepackage[T2A]{fontenc}
\usepackage[russian,english]{babel}
\usepackage{times}
\usepackage{latexsym}
\usepackage{amsmath}
\usepackage[T1]{fontenc}

\usepackage[utf8]{inputenc}

\usepackage{listings}
\usepackage{microtype}
\usepackage{tikz}
\usepackage{caption}
\usepackage{subcaption}
\usepackage{dblfloatfix}
\usepackage{inconsolata}

\usepackage{graphicx}

\usepackage{tikz, verbatimbox, pgfplots}
\usetikzlibrary{positioning,shapes}
\usetikzlibrary {arrows.meta} 
\pgfplotsset{compat=1.18}
\title{MisinfoTeleGraph: Network-driven Misinformation Detection for German Telegram Messages}

\author{
  \textbf{Lu Kalkbrenner\textsuperscript{1}},
  \textbf{Veronika Solopova\textsuperscript{2}},
 \textbf{Steffen Zeiler\textsuperscript{2}},
  \textbf{Robert Nickel\textsuperscript{3}},
\\
  \textbf{Dorothea Kolossa\textsuperscript{2}}
\\
\\
  \textsuperscript{1}CeMAS›,
  \textsuperscript{2}Technische Universität Berlin,
  \textsuperscript{3}Bucknell University
\\
  \small{
    \textbf{Correspondence:} \href{mailto:lu@kalkbrenner.in}{lu@kalkbrenner.in}
  }
}
\begin{document}
\maketitle
\begin{abstract}
 Connectivity and message propagation are central, yet often underutilised, sources of information in misinformation detection—especially on poorly moderated platforms such as Telegram, which has become a critical channel for misinformation dissemination, namely in the German electoral context. In this paper, we introduce Misinfo-TeleGraph, the first German-language Telegram-based graph dataset for misinformation detection. It includes over 5 million messages from public channels, enriched with metadata, channel relationships, and both weak and strong labels. These labels are derived via semantic similarity to fact-checks and news articles using M3-embeddings, as well as manual annotation. To establish reproducible baselines, we evaluate both text-only models and graph neural networks (GNNs) that incorporate message forwarding as a network structure. Our results show that GraphSAGE with LSTM aggregation significantly outperforms text-only baselines in terms of Matthews Correlation Coefficient (MCC) and F1-score. We further evaluate the impact of subscribers, view counts, and automatically versus human-created labels on performance, and highlight both the potential and challenges of weak supervision in this domain. This work provides a reproducible benchmark and open dataset for future research on misinformation detection in German-language Telegram networks and other low-moderation social platforms.
\end{abstract}

\section{Introduction}\label{sec:intro_related_work}

Disinformation and misinformation, with their proven impact on democratic elections, have become one of the most harmful online phenomena of our age \citep{Howard2019IRA}.
Ever since mainstream social media platforms implemented more thorough content moderation policies against harmful speech and misinformation, many users migrated to Telegram \citep{deplatforming_extreme_right}. For instance, it was shown that around 30\% of adults use the Telegram messenger as a news source in Germany \citep{Holnburger_2023_telegram}.
Telegram has become a key platform for spreading misinformation, conspiracy theories and far-right ideologies in Germany, while largely remaining unmoderated \citep{far_right_telegram, Holnburger_2023_telegram}, and solidifying false beliefs with the echo chamber effect \cite{pushshift-traditional-graph-analysis}. 
Already in 2017, the Council of Europe reported that conventional fact-checking was becoming unable to respond to such data volumes to identify check-worthy content and verify it in a timely manner \cite{wardle2017information}. Therefore, in recent years, extensive research has been conducted on identifying misinformation using machine learning methods. However, most studies focused on data from X (formerly Twitter) and on the English language, while for other languages, including German, mostly simple text-based methods were investigated.
In this study, we present our \textbf{Misinfo-TeleGraph} Dataset\footnote{\url{https://zenodo.org/records/13362123}}, which is a German Telegram misinformation graph dataset including 13.845 German Telegram channels and their messages from October 2022 to May 2024, including the forwarding information and metadata regarding views and likes. 742 messages are weakly labeled by corresponding fact-checks and newspaper articles using similarity scores from M3-embeddings. We trained a Graph Neural Network (GNN) to detect misinformation and analyzed how the incorporation of network information improves the model's performance in comparison to a text-only approach. We make our code available in GitHub\footnote{\url{https://github.com/kalkbrennerei/MisinfoTeleGraph}}.

\section{Related Work}
While multiple successful methods were developed to detect the factual correctness of news purely relying on textual content \cite{lstm_content_based, gans_fake_news, FakeBERT, SAFE}, such models were shown to be language-dependent \citep{GCNFN}, prone to adversarial attacks \citep{GNN-CL, adversarial_examples}, and generalize badly to new data due to over-reliance on linguistic patterns and keywords \citep{solopova-etal-2024-check}.
Recent works have been focusing on including social context and propagation patterns \cite{10.1145/3137597.3137600}. Approaches based on social context often focus on user demographics, account authenticity and political bias of the thread participants \citep{user_engagement}, location and profile pictures \cite{user_profile_info}. Other approaches look at social network structure, and user reactions such as likes and shares \citep{GCNFN, li-etal-2020-exploiting, ijcai2020p197}. 
\citet{reply_aided_detection} used message view counts and information about the Telegram channels in which messages have been shared, including the number of subscribers for each channel.

\citet{time_series_propagation_detection} used multivariate time series with recurrent and convolutional networks. \citet{trace_miner_propagation_based} inferred user embeddings with social network structures and classified them using an LSTM-RNN. 
\citet{user_to_user_propagation} analyzed user-to-user interaction propagation paths over multiple hops using a transformer architecture, while \citet{node2vec_fakenews} used node2vec to create graph embeddings from the follower-followee relationship.

Motivated by the graph structure of social networks, \textit{Graph Neural Networks (GNNs)} were identified as a promising technique within propagation-based approaches. 
\citet{GCNFN} applied a GNN for misinformation detection based on data from X, including content, social context, and propagation features. \citet{GNN-CL} extended this approach by leveraging continual learning techniques to improve the performance on unseen data. \citet{UPFD} extracted node features from news articles and user preferences from X using BERT and node2vec embeddings, and compared Graph Convolutional Network (GCN) and GraphSAGE architectures, while also explicitly separating endogenous and exogenous user preferences. Comparing multiple types of GNNs for this task, \citet{comparative-analysis-fakenews} showed that 
GraphSAGE \citep{GraphSAGE} performed best, with a test accuracy of 96.99\%. \citet{MuMiN}, which serves as the main inspiration for our work, implemented a heterogeneous version of the GraphSAGE model as a baseline for their MuMiN dataset of multi-lingual tweets, achieving an F1 score of 61.45\%  compared to the LaBSE (Language-Agnostic BERT Sentence Embedding) text-only baseline of 57.90\%.

Most existing graph-based misinformation detection datasets, like the MuMiN and FakeNewsNet \citep{FakeNewsNet}, are primarily derived from X, with limited options from other social networks. While there are non-specific datasets from platforms like Telegram, such as TGDataset \citep{tgdataset} and the Pushshift dataset \citep{telegram-pushshift}, research on graph neural networks for misinformation detection in Telegram data is notably absent.
While \citet{magnet-telegram-gnn} utilized Telegram threads to train a GNN for a node classification task, to the best of our knowledge, ours is the first work implementing GNNs with Telegram data for misinformation detection, and also the first on employing these for the German language.


\section{Methods}
To create the Telegram graph dataset, we used \textit{weak annotation} on data that we received from \citet{d4t}. From this dataset, we constructed a graph using network information, including messages, channels, views, likes and cross-channel message forwarding. Statistics of the dataset are depicted in Figure~\ref{tab:statistics-d4t-news}. The annotated training graph is used to train a Graph Neural Network, where node embeddings are computed based on their neighborhood representation using the GraphSAGE architecture.

Our methods are inspired by \citet{MuMiN}, who trained a GNN on a graph dataset from X. Since Telegram and X are very different platforms, the creation of our graph dataset differs considerably from the work of \citet{MuMiN}. However, we were able to reuse some of their code, and we employed the same approach to train a baseline GNN model on the data.
\begin{table}[h]
    \centering
    \small
    \begin{tabular}{|l|c|}
        \hline
         \# Telegram channels & 13,845\\
        \hline
       \# Telegram messages & 5,727,631\\
        \hline
        Similarity threshold & 0.7\\
        \hline
       \# weakly linked message-claim pairs & 742\\
        \hline
        \# weak pairs in the factual class & 110\\
        \hline
        \# weak pairs in the misinfo. class & 632\\
        \hline
       \# weak pairs in the `other' class & 542\\
        \hline
        \# strongly linked message-claim pairs &  651\\
        \hline
        \# strong pairs in the factual class & 94\\
        \hline
        \# strong pairs in the misinfo. class & 557\\
        \hline
    \end{tabular}
    \caption[Statistics of the \textit{MisinfoTeleGraph} dataset]{Statistics of the \textit{MisinfoTeleGraph} dataset.}
    \label{tab:statistics-d4t-news}
\end{table}

\begin{figure*}[h!]
    \centering
    \includegraphics[width=0.8\textwidth]{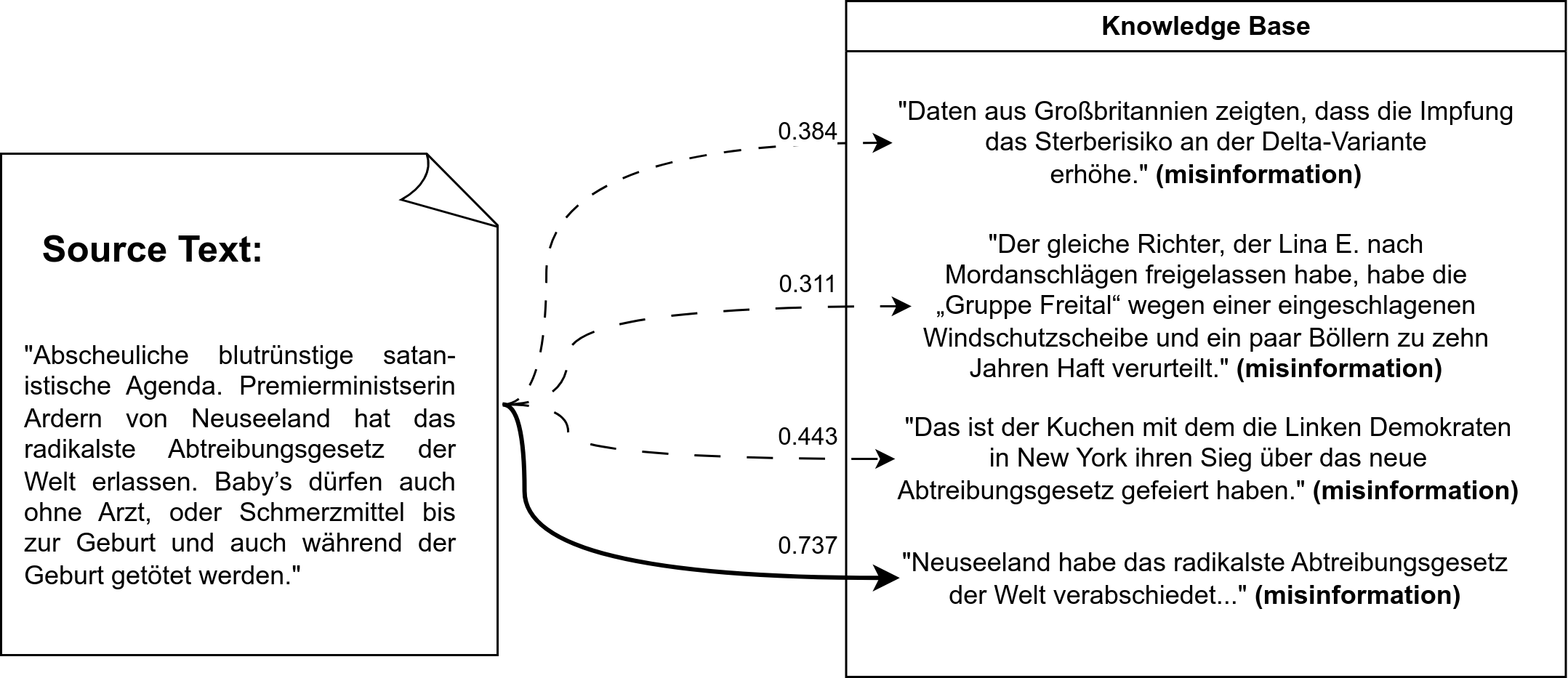}
    \caption{Weak Supervision using M3-embeddings and a knowledge base.
A source text (on the left) is compared to claims contained in a knowledge
base of fact-checks and news articles (on the right). Similarity scores are computed based on the
M3-embeddings of the text and claims. The source text is linked with a claim, if
the similarity score exceeds a threshold of 0.7, as is the case for the last claim
in the knowledge base with a score of 0.737. The source text inherits the label
(factual or misinformation) of the claim that it is matched with.}
    \label{fig:weak-supervision}
\end{figure*}

\subsection{Telegram Data Source}
For training our model, we created a dataset based on data provided from  \citet{d4t} (D4T). 
Their data contains information about which channel messages are posted in and which channel messages are being forwarded to. From this message-forwarding network information, we constructed a graph dataset as described in Section~\ref{s:network-info}.

\subsection{Training Data Annotation}
Since training data annotation remains a costly task, \textit{weak annotation} is a promising approach to annotate data sets of misinformation from online social networks. Manual data annotation often requires skilled human annotators, who are knowledgeable in their domain, such as professional fact-checkers in the case of the detection of misinformation. In this work, we use semantic similarity based on M3-embeddings \cite{bge-m3} to pre-select Telegram messages that potentially contain misinformation and manually annotate the pre-selected collection. This approach is shown in Figure \ref{fig:weak-supervision}.
For the weak annotation, we use a knowledge base of newspaper articles and fact-checking articles that contain texts from the sources in Table \ref{fig:sources}. The fact-checks were fetched from the Google Fact Check Tools API \cite{google-fact-check-api} and the newspaper articles were fetched from WoldNewsAPI \footnote{\href{https://worldnewsapi.com/}{https://worldnewsapi.com/}}.

\begin{table}[h]
    \centering
    \begin{tabular}{|c|c|c|c|}
        \hline
        Source & \# articles\\
        \hline
        BR (Bayrischer Rundfunk) & 343\\
        CORRECTIV & 2568\\
        DPA (Deutsche Presseagentur) & 2271\\
        AFP (Agence France-Presse) & 1012\\
        presseportal.de & 378\\
        \hline
        Zeit & 2396\\
        Taz & 1293\\
        Süddeutsche & 655\\
        \hline
    \end{tabular}
    \caption[German knowledge base sources]{German knowledge base sources. Fact-checking articles on top and newspaper articles below.}
    \label{fig:sources}
\end{table}

The texts from the knowledge base are compared to the telegram messages using semantic similarity. We compared different semantic similarity thresholds by precision. We found that a threshold of 0.7 matches enough message-claim pairs with an acceptable precision of 67.86\%. The resulting 868 weakly annotated message-claim pairs were annotated by hand to obtain 589 strongly annotated message-claim pairs. The precision was computed by dividing the number of strongly annotated message-claim pairs by the number of weakly annotated pairs.

\subsection{Network Information from message forwarding}

\begin{figure*}[h!]
    \centering

        \resizebox{0.6\textwidth}{!}{%
        \begin{tikzpicture}[node distance=3cm, auto, every node/.style={align=center}]
        
        \tikzstyle{message} = [circle, draw, fill=red!20, node distance=3cm, minimum size=2cm]
        \tikzstyle{channel} = [circle, draw, fill=blue!20, node distance=3cm, minimum size=2cm]
        \tikzstyle{partof} = [->, thick, -{Stealth[length=5mm]}]
        \tikzstyle{forwarded} = [->, dashed, -{Stealth[length=5mm]}]
        
        \node[channel] (A) {Q Digital \\ Patrioten};
        \node[message] (B) [right=of A, xshift=2.5cm, yshift=-1cm] {M1};
        \node[message] (C) [below=of B] {M1};
        \node[message] (D) [below=of C] {M1};
        \node[channel] (E) [below left=of C, xshift=-2.5cm, yshift=-1cm] {WAKE UP \\ Media};
        \node[channel] (F) [right=of C, xshift=2.5cm, yshift=-1cm] {Pizzagate \\ Archiv};
        \node[message] (G) [above right=of F, xshift=-2.5cm, yshift=1cm] {M2};
        \node[message] (H) [below right=of F, xshift=-2.5cm, yshift=-1cm] {M3};
        
        \draw[partof] (B) to[bend left] node[pos=0.3, above] {IS\_PART\_OF} (A);
        \draw[forwarded] (B) to[bend right] node[pos=0.7, below] {FORWARDED} (A);
        \draw[partof] (D) to[bend right] node[midway, above] {IS\_PART\_OF} (E);
        \draw[forwarded] (D) to[bend left] node[midway, below] {FORWARDED} (E);
        \draw[partof] (C) to[bend right] node[midway, above] {IS\_PART\_OF} (F);
        \draw[partof] (G) -- node[midway, right] {IS\_PART\_OF} (F);
        \draw[partof] (H) -- node[midway, right] {IS\_PART\_OF} (F);
        \draw[forwarded] (C) -- node[midway, left] {FORWARDED} (A);
        \draw[forwarded] (C) -- node[midway, above] {FORWARDED} (E);
        \draw[forwarded] (F) to[bend right] node[midway, right] {FORWARDED} (C);
        \draw[forwarded] (C) -- node[midway, right] {FORWARDED} (D);
        \draw[forwarded] (C) -- node[midway, below] {FORWARDED} (B);
        
        
        \end{tikzpicture}
        }
    \caption[Social Network Graph of the ``Pizzagate Archiv`` Telegram Channel.]{Example of a Social Network Graph. The example shows the ``Pizzzagate Archiv`` Telegram channel depicted in purple on the right.
    There are three messages (depicted in red) that were posted in the ``Pizzagate Archiv`` channel and are thus connected via an IS\_PART\_OF relation. One of the messages (depicted in the middle) has also been forwarded to two other channels, namely the ``Q Digital Patrioten`` channel and the ``WAKE UP Media`` channel. To preserve the information in which channel a message has been posted first, the messages are duplicated and linked by a FORWARDED relationship to the original messages when they are being forwarded. This is why the message in the middle appears three times in the middle. Every message only has one IS\_PART\_OF relationship with one channel.}
    \label{fig:social-network-graph}
\end{figure*}
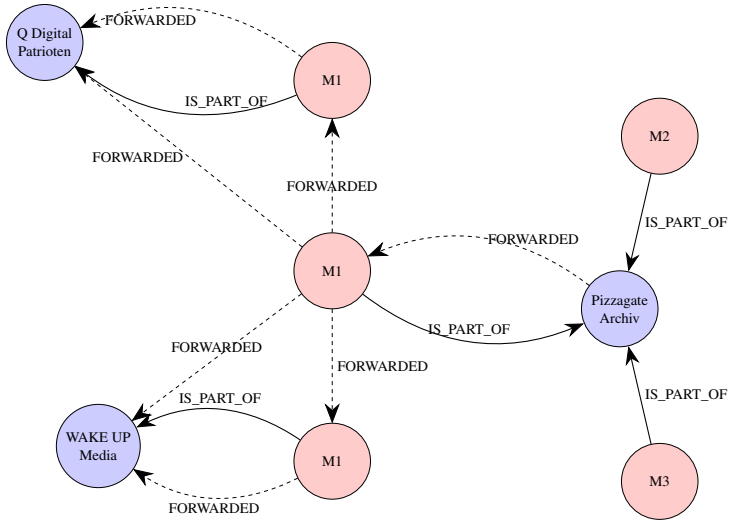

\label{s:network-info}
To feed both textual information and network information into a graph neural network, we created a graph $\mathcal{G}$  with two node classes for Telegram channels and Telegram messages. We use the following two edge classes to describe the information about messages being forwarded across channels:
\begin{itemize}
    \item \verb|IS_PART_OF| describes the relationship of a message being posted in a channel
    \item \verb|FORWARDED| describes the relationship of a message being forwarded to a channel.
\end{itemize}
To preserve the information in which channel a message has been posted first, the messages are duplicated and linked by a \verb|FORWARDED| relationship to the original messages when they are being forwarded.

Every node $n$ has a feature vector $X_n$ that contains the M3-embedding of the message text or the channel description concatenated with metadata.

A subgraph of the graph that we created can be seen in Figure \ref{fig:social-network-graph}.

\subsection{Training of the GNN}
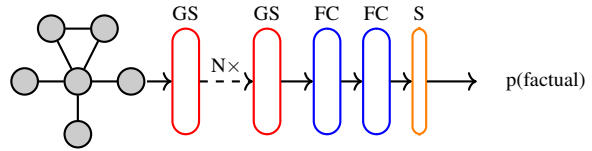
\begin{figure}[h!]
    \centering
    \begin{tikzpicture}[thick, scale=0.7, every node/.style={transform shape}]

        \node[draw, circle, fill=gray!40, minimum size=0.5cm, inner sep=0pt] (a) at (0,0) {};
        \node[draw, circle, fill=gray!40, minimum size=0.5cm, inner sep=0pt] (b) at (-0.5,1) {};
        \node[draw, circle, fill=gray!40, minimum size=0.5cm, inner sep=0pt] (c) at (0.5,1) {};
        \node[draw, circle, fill=gray!40, minimum size=0.5cm, inner sep=0pt] (d) at (-1,0) {};
        \node[draw, circle, fill=gray!40, minimum size=0.5cm, inner sep=0pt] (e) at (1,0) {};
        \node[draw, circle, fill=gray!40, minimum size=0.5cm, inner sep=0pt] (f) at (0,-1) {};
        \draw[thick] (a) -- (b) -- (c) -- (a);
        \draw[thick] (a) -- (d);
        \draw[thick] (a) -- (e);
        \draw[thick] (a) -- (f);
        
        \node[draw, thick, minimum width=0.5cm, minimum height=2cm, right=1.5cm of a, rounded corners=5pt, label=above:GS, red] (gc1) {};
        \node[draw, thick, minimum width=0.5cm, minimum height=2cm, right=1cm of gc1, rounded corners=5pt, label=above:GS, red] (gc2) {};
        
        \node[draw, thick, minimum width=0.5cm, minimum height=2cm, right=0.6cm of gc2, rounded corners=5pt, label=above:FC, blue] (fc1) {};
        \node[draw, thick, minimum width=0.5cm, minimum height=2cm, right=0.4cm of fc1, rounded corners=5pt, label=above:FC, blue] (fc2) {};
        \node[draw, thick, minimum width=0.2cm, minimum height=2cm, right=0.4cm of fc2, rounded corners=3pt, label=above:S, orange] (sm) {};
        
        \node[below] at (8.8,0.32) {p(factual)};
        
        \draw[->, thick] (1.3,0) -- (gc1.west);
        \draw[->, dashed] (gc1.east) -- node[midway, above] {N$\times$}(gc2.west);
        \draw[->, thick] (gc2.east) -- (fc1.west);
        \draw[->, thick] (fc1.east) -- (fc2.west);
        \draw[->, thick] (fc2.east) -- (sm.west);
        \draw[->, thick] (sm.east) -- (7.5,0);

\end{tikzpicture}
    \caption[GraphSAGE architecture.]{GraphSAGE model architecture. The network takes a node to be classified and its surrounding graph as an input. The graph is passed through $N$ GraphSAGE layers (GS). The node embedding of the node to be classified is then passed through two fully connected layers (FC). A Sigmoid function (S) is applied to the resulting logits to compute the probabilities of belonging to the factual or the misinformation class.}
    \label{fig:graph-sage-layers}
\end{figure}

To train the GNN model, we followed the procedure of \cite{MuMiN}, using a GraphSAGE architecture as proposed by \cite{GraphSAGE}.
We experimented with different numbers of GraphSAGE layers and different aggregation functions. The GraphSAGE architecture setup is depicted in Figure \ref{fig:graph-sage-layers}. We set the learning rate to 1e-3 with a learning rate scheduler that starts at 1e-3 and ends at 1e-5 after 100 iterations, using a weight decay of 1e-5 for all experiments.

\section{Experimental Setup}
\label{s:experimental-setup}
For the GNN architecture depicted in Figure \ref{fig:graph-sage-layers}, we experimented with different numbers of GraphSAGE layers, different aggregator architectures and number of epochs.
We then used the best-performing combination to verify our main hypotheses:

\begin{enumerate}
    \item Including additional graph information (forwarding information) to train a GNN has an edge over the text-only misinformation classification baseline.
    \item Including view and subscriber counts improves the GNN baseline.
    \item Using weak labels for GNN training does not result in significantly poorer performance and calibration compared to strong labels.
\end{enumerate}

\begin{figure}[h!]
    \centering
    \begin{tikzpicture}[thick, scale=0.7, every node/.style={transform shape}]

        \draw[thick, fill=gray!10] (0, -1) rectangle (2, 1);  
        \draw[thick, fill=gray!10] (-0.2, -0.8) rectangle (1.8, 1.2);  
        \draw[thick, fill=gray!10] (-0.4, -0.6) rectangle (1.6, 1.4);  
        
        \node at (0.6, 0.5) {Text Data}(a);
        
        \node[draw, thick, minimum width=1cm, minimum height=2cm, right=3cm of a, rounded corners=8pt, label=above:M3, teal] (m3) {};
        
        \node[draw, thick, minimum width=0.6cm, minimum height=2cm, right=0.4cm of m3, rounded corners=5pt, label=above:FC, blue] (fc1) {};
        \node[draw, thick, minimum width=0.2cm, minimum height=2cm, right=0.4cm of fc1, rounded corners=3pt, label=above:S, orange] (sm) {};

        \node[below] at (8.8,0.32) {p(factual)};
        
        \draw[->, thick] (2.1,0) -- (m3.west);
        \draw[->, thick] (m3.east) -- (fc1.west);
        \draw[->, thick] (fc1.east) -- (sm.west);
        \draw[->, thick] (sm.east) -- (7.5,0);

\end{tikzpicture}
    \caption[Text-based model architecture]{Text-based model architecture. Based on the text data of the messages to be classified, M3-embeddings are computed (M3). The embeddings are then classified by a fully connected layer (FC) and a Sigmoid function (S).}
    \label{fig:text-only-layers}
\end{figure}
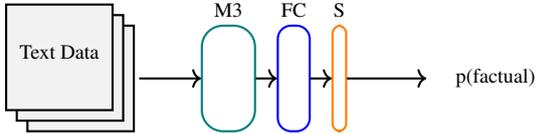

For the text-only baseline, we use an architecture based on M3 that is depicted in Figure \ref{fig:text-only-layers}.

\section{Results}
\subsection{Metrics}

As standard evaluation metrics, we use Precision, Recall and their harmonic average F1-score, considering these indicators separately for misinformation and true samples. We also use the Matthews correlation coefficient (MCC), which is robust to unbalanced datasets, as a combination of precision and recall. Finally, we calculate the Expected Calibration Error (ECE) from \citep{ece}, which measures if a model’s predicted output probabilities reflect the accuracy of its decision, to assess whether the model is exhibiting over-confidence or under-confidence. It is computed by
\begin{equation}
    \text{ECE} = \sum_{b=1}^B \frac{n_b}{N} |\text{acc}(b) - \text{conf}(b)|,
\end{equation}
where $B$ is the number of bins, $n_b$ is the number of predictions in bin $b$, and $N$ is the total number of data points. Each prediction is assigned to a bin based on its confidence score (i.e., the predicted probability of the top class), and $\text{acc}(b)$ and $\text{conf}(b)$ denote the average accuracy and average confidence within bin $b$, respectively.

\subsection{Qualitative findings during annotation}

While manually annotating message–claim pairs generated by the weak annotator model, we observed that it performs surprisingly well in cross-lingual contexts. Despite the dataset being composed of German-language Telegram channels, several English and Russian messages that were also contained in the channels were matched correctly with German claims. For example, an English message about the U.S.~deploying Marines to Israel was successfully paired with a German-language claim falsely reporting that thousands of U.S.~soldiers had landed in Israel (see Table~\ref{tab:cross-lingual}). Further examples can be found in Appendix~\ref{sec:appendix1}.

However, the model often failed to capture logical specificity. For instance, it confused claims about vaccine-related deaths with those referring to COVID-19 fatalities, and did not consistently distinguish between adverse effects and death. Similarly, in messages related to the Gaza conflict, the model was unable to identify which actor—Israel or Hamas—was described as initiating violence.

These cases suggest that, while cross-lingual matching is a strength, the weak annotator model struggles with logical entailment and causal nuance, highlighting a key area for improvement in future work.

\subsection{GNN architecture}
We compare different numbers of GraphSAGE layers, LSTM and mean aggregation, and different numbers of epochs. 

Similar to \citet{MuMiN}, we are able to verify that LSTM aggregation performs best in terms of all considered metrics as depicted in Table~\ref{tab:aggregators-test}. This is likely due to the ability of LSTM to remember long-term dependencies over multiple ``hops" of the graph.

\begin{table}[h]
\centering
\begin{tabular}{|r|c|c|}
    \hline
    & mean agg. & LSTM agg. \\
    \hline
    factual precision & 0.357 & \textbf{0.75}\\
    misinfo precision & 0.935 & \textbf{1.0}\\
    factual recall & 0.833 & \textbf{1.0}\\
    misinfo recall & 0.617 & \textbf{0.914}\\
    factual $F_1$ & 0.5 & \textbf{0.857}\\
    misinfo $F_1$ & 0.744 & \textbf{0.956}\\
    MCC & 0.363 & \textbf{0.828}\\
    \hline
\end{tabular}
\caption[Mean and LSTM aggregator models compared by their metrics]{Test set metrics for mean and LSTM aggregators on the weak training set and 4 GraphSAGE layers after 10 epochs.}
\label{tab:aggregators-test}
\end{table}

Unlike \citet{MuMiN}, which achieved the best performance for two GraphSAGE layers, we are able to observe the best performance in terms of almost all metrics measured for four GraphSAGE layers. Four GraphSAGE layers correspond to four ``hops`` in the graph depicted in Figure~\ref{fig:social-network-graph}. This is likely due to the graph structure, as channel nodes are never directly connected.
\begin{figure*}[htbp!]
    \centering
    \begin{subfigure}[t]{0.48\textwidth}
        \centering
        \begin{tikzpicture}[scale=0.57]
            \begin{axis}[
                width=\linewidth,
                height=7cm,
                xlabel={Number of epochs},
                ylabel={Test Metrics},
                grid=major,
                legend pos=outer north east,
                label style={font=\large},
                legend style={font=\normalsize},
                thick,
            ]
            \addplot[color=teal,mark=*] coordinates {
                (10, 0.9767441749572754)
                (20, 0.95652174949646)
                (30, 0.9130434989929199)
                (50, 0.9791666865348816)
                (70, 0.978723406791687)
            };
            \addlegendentry{Precision misinfo}
            \addplot[color=blue,mark=*] coordinates {
                (10, 0.9438202381134033)
                (20, 0.95652174949646)
                (30, 0.9438202381134033)
                (50, 0.9894737005233765)
                (70, 0.9892473220825195)
            };
            \addlegendentry{F1 misinfo}
            \addplot[color=violet,mark=*] coordinates {
                (10, 0.9130434989929199)
                (20, 0.95652174949646)
                (30, 0.9767441749572754)
                (50, 1.0)
                (70, 1.0)
            };
            \addlegendentry{Recall misinfo}
            \addplot[color=cyan,mark=*] coordinates {
                (10, 0.9230769276618958)
                (20, 0.8461538553237915)
                (30, 0.75)
                (50, 0.9166666865348816)
                (70, 0.9230769276618958)
            };
            \addlegendentry{Recall factual}
            \addplot[color=red,mark=*] coordinates {
                (10, 0.8275862336158752)
                (20, 0.8461538553237915)
                (30, 0.8275862336158752)
                (50, 0.95652174949646)
                (70, 0.9599999785423279)
            };
            \addlegendentry{F1 factual}
            \addplot[color=green,mark=*] coordinates {
                (10, 0.75)
                (20, 0.8461538553237915)
                (30, 0.9230769276618958)
                (50, 1.0)
                (70, 1.0)
            };
            \addlegendentry{Precision factual}
            \addplot[color=orange,mark=*] coordinates {
                (10, 0.7795162796974182)
                (20, 0.8026756048202515)
                (30, 0.7795162796974182)
                (50, 0.9474014639854431)
                (70, 0.9504930377006531)
            };
            \addlegendentry{MCC}
            \end{axis}
        \end{tikzpicture}
        \caption{Varying the number of epochs.}
        \label{fig:epochs-subplot}
    \end{subfigure}
    \hfill
    \begin{subfigure}[t]{0.48\textwidth}
        \centering
        \begin{tikzpicture}[scale=0.57]
            \begin{axis}[
                width=\linewidth,
                height=7cm,
                xlabel={Number of GraphSAGE Layers},
                ylabel={Test Metrics},
                xtick={0,1,...,4},
                grid=major,
                legend pos=outer north east,
                label style={font=\large},
                legend style={font=\normalsize},
                thick,
            ]
            \addplot[color=teal,mark=*] coordinates {
                (1, 0.914)
                (2, 0.935)
                (3, 0.911)
                (4, 0.928)
            };
            \addlegendentry{Precision misinfo}
            \addplot[color=blue,mark=*] coordinates {
                (1, 0.780)
                (2, 0.743)
                (3, 0.891)
                (4, 0.876)
            };
            \addlegendentry{F1 misinfo}
            \addplot[color=violet,mark=*] coordinates {
                (1, 0.680)
                (2, 0.617)
                (3, 0.872)
                (4, 0.829)
            };
            \addlegendentry{Recall misinfo}
            \addplot[color=cyan,mark=*] coordinates {
                (1, 0.75)
                (2, 0.833)
                (3, 0.666)
                (4, 0.75)
            };
            \addlegendentry{Recall factual}
            \addplot[color=red,mark=*] coordinates {
                (1, 0.5)
                (2, 0.5)
                (3, 0.615)
                (4, 0.620)
            };
            \addlegendentry{F1 factual}
            \addplot[color=green,mark=*] coordinates {
                (1, 0.375)
                (2, 0.357)
                (3, 0.571)
                (4, 0.529)
            };
            \addlegendentry{Precision factual}
            \addplot[color=orange,mark=*] coordinates {
                (1, 0.353)
                (2, 0.363)
                (3, 0.509)
                (4, 0.515)
            };
            \addlegendentry{MCC}
            \end{axis}
        \end{tikzpicture}
        \caption{Varying the number of GraphSAGE layers.}
        \label{fig:layers-subplot}
    \end{subfigure}
    \caption[Test set metrics for different training configurations]{Test set metrics across two training configurations: (a) different numbers of training epochs using LSTM aggregators and 4 GraphSAGE layers.  
after 100 iterations, and a weight decay of 1e-5; (b) different numbers of GraphSAGE layers using mean aggregation. 10 epochs, learning rate 1e-3 with the same learning
rate scheduler.}
    \label{fig:test-metrics-subplots}
\end{figure*}
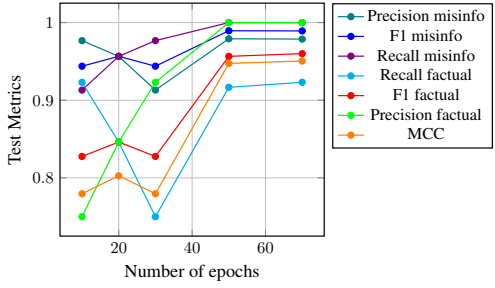
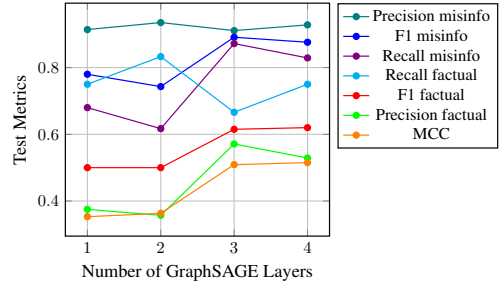
The four-hop neighborhood of a Telegram message contains all messages of the same channel and all messages of the channels they are being forwarded to.
We were unable to test more than four GraphSAGE layers due to hardware restrictions.
Figure \ref{fig:test-metrics-subplots} depicts the results for different numbers of GraphSAGE layers and numbers of epochs.

Hence, the overall best-performing architecture is the one using LSTM aggregation and 4 GraphSAGE layers. Due to hardware restrictions, we use 10 epochs for the experiments in the following sections.

\subsection{Comparison of GNN and Text-only Model}
\label{s:comparison-text-model}

\begin{table}[h]
\centering
\begin{tabular}{|r|c|c|}
    \hline
    & graph-based & text-based \\
    \hline
    factual precision & \textbf{1.0} & 0.714\\
    misinfo precision & \textbf{0.979} & 0.943\\
    factual recall & \textbf{0.923} & 0.833\\
    misinfo recall & \textbf{1.0} & 0.893\\
    factual $F_1$ & \textbf{0.960} & 0.769\\
    misinfo $F_1$ & \textbf{0.989} & 0.917\\
    MCC & \textbf{0.950} & 0.691\\
    \hline
\end{tabular}
\caption[Text-only baseline metrics in comparison with the GNN baseline]{Test set metrics for the text-only baseline in comparison with the graph baseline. The graph baseline uses an LSTM aggregator and 4 GraphSAGE layers.}
\label{tab:text-vs-graph}
\end{table}
In this Section, we compare the GNN model depicted in Figure~\ref{fig:graph-sage-layers} to the text-based model depicted in Figure~\ref{fig:text-only-layers}. Table~\ref{tab:text-vs-graph} shows the different metrics for the two baselines. The GNN model outperforms the text-based model for all metrics. We achieved a 95\% MCC score for the graph-based model and 69.1\% MCC for the text-based model. Our results are comparable to those of \citet{comparative-analysis-fakenews}, who achieve 78.12\% test accuracy for a text-based model and 96.99\% test accuracy for a GraphSAGE model for the classification of misinformation.
\vspace{-0.2em}
The results from this section verify our hypothesis (1) from Section~\ref{s:experimental-setup} that taking additional network information into account improves performance over the text-only misinformation classification baseline.


\subsection{Effect of using View and Subscriber Counts}
\label{s:view-subscriber}

\begin{table}[h!]
\centering
\begin{tabular}{|r|c|c|}
    \hline
    & incl. counts & w/o counts \\
    \hline
    factual precision & \textbf{1.0} & 0.923\\
    misinfo precision & \textbf{0.979} & 0.978 \\
    factual recall & \textbf{0.923} & \textbf{0.923}\\
    misinfo recall & \textbf{1.0} & 0.978\\
    factual $F_1$ & \textbf{0.960} & 0.923\\
    misinfo $F_1$ & \textbf{0.989} & 0.978\\
    MCC & \textbf{0.950} & 0.901\\
    \hline
\end{tabular}
\caption[GNN applied to the datasets including subscriber and view counts and without including them compared by their test metrics]{GNN applied to the datasets including subscriber and view counts and without including them compared by their metrics. The model uses an LSTM aggregator and 4 GraphSAGE layers.}
\label{tab:view-subscriber}
\end{table}

\begin{table}[h]
\centering
\begin{tabular}{|r|c|c|}
    \hline
    & weak data & strong data \\
    \hline
    factual precision & \textbf{1.0} & \textbf{1.0}\\
    misinfo precision & \textbf{0.979} & 0.974 \\
    factual recall & \textbf{0.923} & 0.875\\
    misinfo recall & \textbf{1.0} & \textbf{1.0}\\
    factual $F_1$ & \textbf{0.960} & 0.933\\
    misinfo $F_1$ & \textbf{0.989} & 0.987\\
    MCC & \textbf{0.950} & 0.923\\
    ECE & \textbf{0.033} & 0.051\\
    \hline
\end{tabular}
\caption[GNN applied to the weak and strong datasets compared by their metrics]{GNNs trained on the weak and strong datasets, compared by their test metrics. The model uses an LSTM aggregator and 4 GraphSAGE layers.}
\label{tab:weak-vs-strong-test}
\end{table}
\begin{table*}[h!]
\centering
\begin{tabular}{|l|r|r|r|}
\hline
\textbf{Channel} & $C_{D_f}$ & \textbf{Out} & \textbf{In} \\
\hline
Eva Herman Offiziell & 17,522 & 16,420 & 1,102 \\
Tagesereignisse der Offenbarung & 13,617 & 1,084 & 12,533 \\
AUF1 & 12,969 & 12,966 & 3 \\
Impfen-nein-danke.de & 11,424 & 437 & 10,987 \\
Freie Sachsen & 11,290 & 11,157 & 133 \\
\hline
\end{tabular}
\caption{Top 5 channels by forward-degree centrality ($C_{D_f}$)}
\label{tab:forward-centrality-top5}
\end{table*}

To compute the node features used in the GNN model in all previous experiments, we concatenated the M3-embedding with additional metadata. 
The embedding of the channel name was concatenated with the number of subscribers. The message embedding was concatenated with the number of views. In this Section, we removed the view and subscriber counts to test if the model performs worse. The results can be seen in Table~\ref{tab:view-subscriber}. The model that uses only M3-embeddings and does not have access to view and subscriber counts on the right performs slightly worse for all metrics except factual recall. This verifies the hypothesis (2) from Section~\ref{s:experimental-setup}.

\subsection{Weak and Strong Labels}
\label{s:comparison-weak-strong}
In this section, we compare the weak and strong datasets. Weakly annotated datasets introduce some noise because there are training examples that have incorrect labels. In some cases, this may lead to perturbations of the classifier \citep{weak-supervision-ranking}, but in many cases, the results are still promising \citep{covid-tweet-annotation-llms}. To test if the weak dataset perturbs the classifier, we manually annotated the test set. Table \ref{tab:weak-vs-strong-test} shows a comparison of the metrics over the weak and strong datasets. The classifier performs similarly on both datasets, which suggests that there are no strong perturbations, verifying our hypothesis (3) from Section~\ref{s:experimental-setup}. 

In this experiment, we also computed the ECE for both classifiers, trained on the weak and the strong dataset. Both values are below 0.1, which implies that both models are calibrated well.
It remains to be seen in future work if we can confirm these results with a
 larger weakly-annotated dataset.

\section{Additional network analysis}
\label{s:graph-analysis-brief}
To illustrate the potential of the \textit{MisinfoTeleGraph} dataset for network analysis, we explore structural properties of the message forwarding graph. The dataset includes forwarding relations between Telegram messages and channels, allowing for classic social network analysis such as centrality computations.

Inspired by 
\citet{centrality-survery-2018} and  \citet{centrality-measures-social-networks}, we computed several centrality measures using the Neo4j Graph Data Science (GDS)\footnote{https://github.com/neo4j/graph-data-science} library, including variations of \textit{degree centrality} and \textit{betweenness centrality}. These measures highlight the most influential Telegram channels in terms of content dissemination and information flow.
Since degree centrality takes all edges into account, we introduce a variant of degree centrality, which we named \textit{forward-degree centrality}. This metric specifically counts the number of edges that represent content forwarding actions. Unlike traditional degree centrality, which includes all edge classes, forward-degree centrality captures only the edges from the \verb|FORWARDED| class. This measure allows to capture information propagation across the platform, reflecting how actively a Telegram channel participates in origination and redistribution patterns of misinformation-related messages.\\
Table~\ref{tab:forward-centrality-top5} shows the top-ranked channels according to this measure. We also differentiate between indegree and outdegree as is usually done for degree centrality. 

Notably, channels like \textit{AUF1} and \textit{Freie Sachsen} act as \textit{broadcast hubs} with high outgoing edge counts, while others like \textit{Tagesereignisse der Offenbarung} mostly redistribute external content. This asymmetry illustrates distinct roles in the misinformation ecosystem — original content creators versus amplifiers — and offers interpretable context for GNN-based classification. Additional centrality metrics, extended tables, and Cypher queries are available in Appendix~\ref{sec:appendix}.


\section{Discussion}

The evaluation of our GNN-based misinformation detection model on the MisinfoTeleGraph dataset has yielded several key insights.

Quantitative evaluation showed that our GNN-based model outperformed a text-only baseline. The graph-based approach achieved an MCC of 0.95 compared to 0.69 for the text-only model, confirming that incorporating network structure improves misinformation classification. Additional experiments with different numbers of GraphSAGE layers indicated that four layers provided the best performance, likely due to the specific Telegram message forwarding network structure. Moreover, the use of an LSTM aggregator consistently outperformed mean aggregation, underscoring the importance of long-term dependency capture in graph-based misinformation detection.

Additionally, we identified cross-lingual capacities of the chosen embeddings, successfully matching German claims with messages in English and Russian on multiple samples. However, qualitative evaluation revealed limitations in handling logical entailment, particularly in differentiating specific statistical claims related to COVID-19 and distinguishing actors in conflicts like the Gaza war. Finally, we noticed that statistical claims that are often found in health-related misinformation, this topic remains hard to classify.

\section{Conclusion}

This study shows that integrating network information into misinformation detection models improves performance over text-only approaches. We present the MisinfoTeleGraph dataset and a reproducible baseline to support future research. Our findings highlight AI's potential in fact-checking, while acknowledging its limits in logical entailment and bias.

AI should assist, not replace, human verification, especially as its generative power still outpaces its detection, reinforcing the need for media literacy and broader misinformation countermeasures.  Future work should focus on multi-modal detection, better weak annotations, and ethical deployment in sensitive contexts to build more robust misinformation detection systems.
\section*{Limitations}
One of the main limitations of this study is the relatively small dataset size. The weakly annotated dataset contains 873 message-claim pairs, and the strongly annotated dataset consists of only 651 pairs. The small dataset size may contribute to potential overfitting and could lead to inflated performance metrics. Future work should aim to scale up the dataset by increasing the number of similarity scores computed per claim and exploring additional sources for annotation. Additionally, misinformation often spreads through multi-modal content such as images and videos, which were not considered in this study. Integrating multi-modal features into the dataset could further improve misinformation detection models.

Another issue relates to data redundancy. During annotation, many messages were found to be thematically similar due to message forwarding and minor text modifications. This raises concerns about potential data leakage, where similar messages may appear in both the training and test sets. Implementing stricter data-splitting techniques, such as clustering similar messages before partitioning, could help mitigate this risk.

The weak annotation approach used in this study was computationally intensive and thereby renders a future expansion of the dataset difficult. The current method relies on computing similarity scores between messages and claims using M3-embeddings, which is effective but slow. Future research should explore hybrid retrieval methods, such as combining BM25 for fast pre-selection with M3-embeddings for precise matching. While GraphSAGE was effective in capturing network structures, alternative GNN architectures could further enhance performance. Graph Attention Networks (GAT) or Graph Isomorphic Networks (GIN) may provide improvements by learning more complex interactions within the network. Additionally, techniques such as neighborhood extension via k-nearest neighbors could help address issues related to low-degree nodes, ensuring that nodes with fewer connections still receive sufficient contextual information.

Finally, deploying a GNN-based misinformation detection model in real-world settings presents challenges due to the need for network information. Unlike text-based models that require only message input, GNNs rely on the surrounding network structure. To facilitate deployment, a continuously updated graph database representing the Telegram ecosystem would be necessary. However, integrating the model into an online fact-checking system or browser extension could provide users with real-time misinformation alerts and be used for selecting check-worthy occurrences for fact-checkers to consider. It would be especially valuable to create cross-lingual and cross-platform graphs to identify coordinated campaigns across languages and different social media websites.
\section*{Acknowledgement}
The work on this paper was mainly performed in the scope of the “noFake” project funded by the German Federal Ministry of Research, Technology and Space (BMFTR) under
Award Identifier F16KIS1519, while the first author was still affiliated with TU Berlin.The manuscript was completed under “news-polygraph” project (BMFTR, reference: 03RU2U151C).

\bibliographystyle{acl_natbib}

\appendix
\section{Examples for Cross-lingual Message-claim Pairs}
\label{sec:appendix1}
\begin{table}[!htbp]
\centering
\begin{adjustbox}{width=\columnwidth}
\begin{tabular}{|p{.48\columnwidth} | p{.48\columnwidth}|}
\hline
\textbf{Message} & \textbf{Claim} \\
\hline
 The United States is sending 2 thousand marines from the rapid reaction brigade to the shores of Israel... & Tausende von US-Marines oder Soldaten sind gerade in Israel gelandet (misinformation – DPA). \\
\hline
NEW – Large German health insurance company analyzed data from 10.9 million insured individuals regarding vaccination complications. "According to our calculations, we consider 400,000 visits to the doctor by our policyholders because of vaccination complications... & Bei der Techniker Krankenkasse seien im Jahr 2021 knapp 440.000 Fälle von Impfnebenwirkungen erfasst worden. In Blog-Artikeln werden die Zahlen mit Werten für 2019 und 2020 verglichen und mit Impfschäden in Verbindung gebracht. (misinformation – CORRECTIV) \\
\hline
NEW – U.S. CDC has quietly deleted the statement that the "mRNA and the spike protein do not last long in the body" from their website... & US-Behörde CDC gibt, dass mRNA und Spikeprotein lange im Körper verbleiben und löscht Entwarnung zu Corona-Impfstoffen von ihrer Webseite. (misinformation – DPA) \\
\hline
\begin{otherlanguage}{russian}Экономика России приходит в упадок – Путин загоняет свою страну в пропасть   \end{otherlanguage} (“Russia's economy is in decline – Putin is driving his country into the abyss” – DeepL translation) & Russlands Kriegswirtschaft: Putin ruiniert sein Land (newspaper article – taz) \\
\hline
\end{tabular}
\end{adjustbox}
\caption{Cross-lingual message–claim pairs}
\label{tab:cross-lingual}
\end{table}
\section{Graph Network Analysis}
\label{sec:appendix}

In this appendix, we propose exemplary graph network analyses that can be done using the MisinfoTeleGraph dataset.

\begin{table}[!htbp]
\centering
\resizebox{\columnwidth}{!}{%
\begin{tabular}{|l|r|r|}
\hline
\textbf{Channel} & \textbf{Subscribers} & \textbf{Degree Centrality} \\
\hline
AUF1 & 252,897 & 12,969 \\
Eva Herman Offiziell & 185,259 & 34,835 \\
Freie Sachsen & 148,628 & 11,290 \\
Tagesereignisse der Offenbarung & 2,045 & 46,925 \\
WELT & 547 & 86,962 \\
\hline
\end{tabular}
}
\caption{Top 5 Telegram channels by subscriber count and degree centrality}
\label{tab:top-channels-summary}
\end{table}

\subsection{Degree Centrality Analysis}
\label{s:centrality-analysis}

\textit{Degree Centrality} captures the immediate influence of a node by counting its direct connections \cite{centrality-survery-2018}. Formally, it is defined as
\begin{equation}
C_D(x) = d_x
\end{equation}
where $d_x$ is the degree of the node.
It is one of the centrality measures that is the easiest to compute in $\mathcal{O}(n)$ time because the algorithm iterates over every node once and counts the number of nodes to which the node is linked.

The following cypher query was used to compute the Degree Centralities.
\begin{lstlisting}[caption={Cypher query for computing Degree Centrality}]
CALL gds.degree.write(
  'messages_and_channels',
  { writeProperty: 'degree' }
) YIELD centralityDistribution, nodePropertiesWritten
RETURN centralityDistribution.min as minScore, 
       centralityDistribution.mean as meanScore, 
       nodePropertiesWritten
\end{lstlisting}

Running the degree centrality query took 41992 ms. The minimum score was 0, which means that there are isolated nodes that do not have neighbors, and the mean score was 2.96. Since a Telegram message can only be part of one Telegram channel, a high degree means that a Telegram message has been forwarded many times. 
\begin{table}[!htbp]
\centering
\begin{adjustbox}{width=\columnwidth}
\begin{tabular}{|p{.11\columnwidth} | p{.88\columnwidth} | p{.25\columnwidth} |}
\hline
\textbf{$C_D$} & \textbf{Channel} & \textbf{Subscribers} \\
\hline
86962 & WELT & 547 \\ \hline
46925 & Tagesereignisse der Offenbarung & 2045 \\ \hline
46827 & impfen-nein-danke.de offiziell  & 11093 \\ \hline
45146 & BILD & 492 \\ \hline
39649 & OutoftheBoxTV\_DerIrrsinnhatProgramm & 4548 \\ \hline
35872 & Schuberts Lagemeldung - Stefan Schubert Offiziell & 36247 \\ \hline
34835 & Eva Herman Offiziell & 185259 \\ \hline
33814 & Aufgewacht & 75 \\ \hline
32844 & Alternative News  & 694 \\ \hline
31728 &  Nyx News | Ukraine  & 1123 \\ \hline
\end{tabular}
\end{adjustbox}
\caption{Top 10 Degree Centrality ($C_D$) of Telegram channels}
\label{tab:degree-centrality-10channels}
\end{table}

Table \ref{tab:degree-centrality-10channels} depicts the ten Telegram channels with the highest degrees, where the ``WELT`` channel is by far the channel with the highest degree.
Note that there are different edge types that channels have that are counted here. They are ingoing and outgoing \verb|FORWARDED| edges and \verb|IS_PART_OF| edges and not all Telegram messages for each channel are included in the dataset. That means that the Degree Centrality of a Telegram channel can be interpreted as how influential a Telegram channel is according to the topics in the fact-checking and news articles.
\subsection{Forward-Degree Centrality Analysis}
To determine influential Telegram channels, the number of how many messages forwarded from or to them, the \verb|FORWARDED| edge type is an interesting feature. Therefore, we computed a new measure, forward-degree centrality, which is defined as
\begin{equation}
C_{D_f}(x) = d_{x_f}
\end{equation}
where $d_{x_f}$ is the degree of $x$ when only taking the \verb|FORWARDED| edge type into account. The measure can be interpreted as how influential a Telegram channel is in spreading information that is related to the fact-checking articles.
We used the following Cypher projection where the \verb|IS_PART_OF| relations are dropped, and computed the degree centrality on the graph projection:
\begin{lstlisting}[caption={Cypher query for computing forward-degree centrality}]
CALL gds.graph.create.cypher(
    'messages_and_channels_forwards',
    'MATCH (n) where (n:TGMessage and n.degree > 1) or n:TGChannel 
    RETURN id(n) AS id',
    'MATCH (n)-[e:IS_FORWARDED_FROM | IS_FORWARDED_TO]-(m) 
    RETURN id(n) AS source, e.weight AS weight, id(m) AS target')

\end{lstlisting}

\subsection{Ingoing and outgoing edges}
 The number of outgoing and ingoing edges can vary a lot in some cases, as can be seen in Table \ref{tab:forward-degree-centrality-channels10}, where many channels have only a few ingoing edges, but a lot of outgoing edges. This might indicate that they create a lot of content that gets frequently forwarded, but do not typically forward messages from other channels themselves.
On the other side, e.g. ``Impfen Nein Danke``, ``Tagesereignisse der Offenbarung`` are channels that often forward information but do not create new original content.
\subsection{Betweenness centrality}
\textit{Betweenness centrality} is a measure that determines the actor that controls information among other nodes via connecting paths \cite{centrality-survery-2018}.

The Betweenness centrality $C_B(x)$ of a node $x$ is defined by
\begin{equation}
C_B(x) = \sum_{u \neq v \in \mathcal{V}(G)} \frac{\sigma_{uv}(x)}{\sigma_{uv}}
\end{equation}
where $\sigma_{uv}$ is the number of shortest $u-v$ paths and $\sigma_{uv}(x)$ is the number of shortest $u-v$ paths that contain $x$.
Computing the Betweenness centrality for a graph with $n$ nodes and $m$ edges has a time complexity of $\mathcal{O}(nm)$ \cite{centrality-survery-2018}. For the graph created from our data, computing the Betweenness scores took around 2 months.

The following cypher query was used to compute the Betweenness Centralities:
\begin{lstlisting}[caption={Cypher query for computing Betweenness Centralities}]
CALL gds.betweenness.write('messages_and_channels',
    { writeProperty: 'betweenness' })
    YIELD centralityDistribution, nodePropertiesWritten
    RETURN centralityDistribution.min AS minimumScore, 
    centralityDistribution.mean AS meanScore, nodePropertiesWritten

\end{lstlisting}

The results are depicted in Table \ref{tab:betweenness-centrality-channels}. The ten Telegram channels with the highest Betweenness centrality are either part of the 10 channels with the highest degree centrality or with the highest forward-degree centrality. 

\begin{table}[!htbp]
\centering
\begin{adjustbox}{width=\columnwidth}
\begin{tabular}{|p{.38\columnwidth} | p{.47\columnwidth} | p{.25\columnwidth} |}
\hline
\textbf{$C_B$} & \textbf{Channel} & \textbf{Subscribers} \\
\hline
1815135347746 & Tagesereignisse der Offenbarung & 2045 \\ \hline
1540030673515 & AUF1 & 252897 \\ \hline
1401343526977 & Eva Herman Offiziell & 185259 \\ \hline
1344917542040 & Aufgewacht & 75 \\ \hline
988463077140 & WELT & 547 \\ \hline
914524232655 & Freie Sachsen & 148628 \\ \hline
811428932945 & impfen-nein-danke.de offiziell & 11093 \\ \hline
810381539513 & henning rosenbusch - channel & 65474 \\ \hline
709972834297 & Mäckle macht gute Laune & 130755 \\ \hline
689059267090 & OutoftheBoxTV & 4548 \\ \hline
\end{tabular}
\end{adjustbox}
\end{table}
\begin{table*}[!htbp]
\caption{Top 10 Forward-Degree Centrality ($C_{D_f}$) of Telegram channels}
\label{tab:forward-degree-centrality-channels10}
\centering
\begin{tabular}{|p{.10\textwidth} | p{.08\textwidth}| p{.10\textwidth} | p{.45\textwidth} | p{.15\textwidth} |}
\hline
\textbf{$C_{D_f}$} & \textbf{Out} & \textbf{In} & \textbf{Channel} & \textbf{Subscribers} \\
\hline
17522 & 16420 & 1102 & Eva Herman Offiziell & 185259 \\ \hline
13617 & 1084 & 12533 & Tagesereignisse der Offenbarung & 2045 \\ \hline
12969 & 12966 & 3 & AUF1 & 252897 \\ \hline
11424 & 437 & 10987 & impfen-nein-danke.de offiziell  & 11093 \\ \hline
11290 & 11157 & 133 & Freie Sachsen & 148628 \\ \hline
10139 & 7537 & 2602 & Mäckle macht gute Laune & 130755 \\ \hline
10192 & 30 & 10162 & OutoftheBoxTV\_DerIrrsinnhatProgramm & 4548 \\ \hline
9576 & 9549 & 27 & henning rosenbusch - channel & 65474 \\ \hline
8575 & 7499 & 1076 & Haintz.Media  \#FreeAssange & 81527 \\ \hline
8135 & 511 & 7624 & RBK - Ceterum censeo NATO esse delendam! Raus aus der NATO! & 2045 \\ \hline
\end{tabular}
\caption{Top 10 Betweenness Centrality ($C_B$) of Telegram channels}
\label{tab:betweenness-centrality-channels}
\end{table*}
\end{document}